\documentclass[letterpaper, 10 pt, conference]{ieeeconf}
\IEEEoverridecommandlockouts                              
\DeclareUnicodeCharacter{0301}{\'{ }}
\usepackage{float} 
\usepackage{tabu}
\usepackage{mathtools}
\usepackage{hyperref}
\usepackage{amssymb}
\usepackage{bbm}
\usepackage{graphicx}
\usepackage{multicol}
\usepackage{multirow}
\usepackage[font=small,labelfont=bf]{caption}
\usepackage{subcaption}
\usepackage{algorithmic}
\usepackage{algorithm}
\let\labelindent\relax
\usepackage{enumitem,kantlipsum}
\usepackage{tablefootnote}
\usepackage{booktabs}
\usepackage{dsfont}
\usepackage{gensymb}
\usepackage{wrapfig}
\usepackage{capt-of}
\usepackage[backend=biber,
            hyperref=true,
            url=false,
            isbn=false,
            doi=false,
            backref=false,
            style=ieee,
            natbib=true,
            citestyle=numeric-comp,
            sorting=none,
            block=none]{biblatex}
\renewcommand{\bibfont}{\small}
\let\ACMmaketitle=\maketitle
\renewcommand{\maketitle}{\begingroup\let\footnote=\thanks \ACMmaketitle\endgroup}
\makeatletter
\usepackage{graphicx}
\let\@fnsymbol\@arabic
\makeatother
\newcommand{\abababa}{~\cite{tremblay2018deep,ha2022flingbot,rao2020rl,anderson2021sim,RetinaGAN,narang2022factory}}
\newcommand{\ababaab}{~\cite{rao2020rl,james2019sim,ho2021retinagan,yuan2022sim}}
\newcommand{\abab}{~\cite{tobin2017domain,tobin2018domain,ren2019domain,peng2018sim,du2021auto}}
\newcommand{\fdsfdsfd}{~\cite{varleyshape,mahler2016dex,mahler2017dex,mahler2018dex,mahler2019learning,viereck2017learning}}
\newcommand{\lkjhg}{~\cite{BENBRAHIM1997283,Kohl2004PolicyGR,Tedrake2004StochasticPG}}
\newcommand{\jkghkg}{~\cite{GULLAPALLI1995237,PETERS2008682,Deisenroth-RSS-11}}
\newcommand{\ikjhuj}{~\cite{wu2020spatial,sun2022fully,gupta2018robot}}
\newcommand{\iiuhnjmk}{~\cite{saycan2022arxiv,huang2022inner}}
\newcommand{\ikyhtg}{~\cite{xia2021relmogen,li2020hrl4in,hadjivelichkov2021improved}}
\newcommand{\rtyhuj}{~\cite{wu2020spatial,sun2022fully,gupta2018robot,saycan2022arxiv,huang2022inner,zhang2022visually,jauhri2022robot,fu2022learning,mittal2021articulated,wang2020learning}}
\newcommand{\tyghnm}{~\cite{tremblay2018deep,ha2022flingbot,varleyshape,mahler2016dex,mahler2017dex,mahler2018dex,mahler2019learning,chebotar2019closing}}
\newcommand{\pepepe}{~\cite{doi:10.1126/scirobotics.abk2822,kumar2021rma,fu2022minimizing,fu2022coupling,miki2022learning,fankhauser2018robust,lee2020learning}}
\newcommand{\okjnhgt}{~\cite{tan2018sim,BipedLocomotion,allevato2020tunenet,david2020iterative,xie2020learning}}
\newcommand{\sdfghjk}{\pepepe\okjnhgt}
\newcommand{\ididid}{~\cite{anderson2021sim,tai2017virtual,kulhanek2019vision,hu2021sim,shen2019situational,zhu2017target}}
\newcommand{\mfrl}{~\cite{Pinto2016SupersizingSL,kalashnikov2018qt,zeng2018learning,Ghadirzadeh2017DeepPP,nair2018visual,singh2019end,Kalashnikov2021MTOptCM}}
\newcommand{\mbrl}{~\cite{pmlr-v97-hafner19a,hafner2019dream,finn2016unsupervised,finn2017deep,ebert2018visual,ebert2018robustness,Lin2019ExperienceEmbeddedVF,Suh2020TheSE,tian2020model,bhardwaj2020information,rafailov2020offline,wu2022daydreamer}}
\newcommand{\ahi}{\ididid}
\newcommand{\member}{\operatorname{\text{M-EMBER}}}
\newcommand{\mobilemember}{\operatorname{\text{Mobile-EMBER}}}
\newcommand{\ember}{\operatorname{\text{EMBER}}}
\newcommand{\enc}{f^k_\text{enc}}

\newcommand{\vae}{f^k_\text{vae}}
\newcommand{\rew}{f^k_{\mathcal{R}}}
\newcommand{\q}{f^k_{Q}}
\newcommand{\qtarget}{f^k_{Q_{target}}}
\newcommand{\ckck}{\texttt{cleaning\_kitchen}}

\begin{document}
\title{\LARGE \bf $\member$: Tackling Long-Horizon Mobile Manipulation via Factorized Domain Transfer}
\author{Bohan Wu\\
Department of Computer Science\\
Stanford University\\
Stanford, CA, USA \\
\texttt{bohanwu@stanford.edu}
\and
Roberto Mart\'{i}n-Mart\'{i}n\\
Department of Computer Science \\
The University of Texas at Austin\\
Austin, TX, USA \\
\texttt{robertomm@utexas.edu}
\and
Li Fei-Fei\\
Department of Computer Science \\
Stanford University\\
Stanford, CA, USA \\
\texttt{feifeili@cs.stanford.edu}
}

\maketitle

\begin{abstract}
In this paper, we propose a method to create visuomotor mobile manipulation solutions for long-horizon activities. We propose to leverage the recent advances in simulation to train visual solutions for mobile manipulation. While previous works have shown success applying this procedure to autonomous visual navigation and stationary manipulation, applying it to long-horizon visuomotor mobile manipulation is still an open challenge that demands both perceptual and compositional generalization of multiple skills. In this work, we develop $\mobilemember$, or $\member$, a factorized method that decomposes a long-horizon mobile manipulation activity into a repertoire of primitive visual skills, reinforcement-learns each skill, and composes these skills to a long-horizon mobile manipulation activity. On a mobile manipulation robot, we find that $\member$ completes a long-horizon mobile manipulation activity, \texttt{cleaning\_kitchen}, achieving a 53\% success rate. This requires successfully planning and executing five factorized, learned visual skills.
\end{abstract}

\section{Introduction}
Mobile manipulators, robots combining locomotion and interaction capabilities, have the potential to undertake multiple long-horizon activities in human environments. Different from short-horizon stationary manipulation such as pushing or grasping, long-horizon mobile manipulation activities require the correct combination of multiple sensorimotor skills to be accomplished. 
Moreover, given the large variability in human environments combined with the challenge of moving the base between interactions, mobile manipulation solutions for the real world have additional demands in generalization and ask for new approaches to learning general visuomotor solutions.

To acquire generalized visuomotor behaviors for \textit{stationary manipulation} in the real world, the robot learning community has resorted to two main procedures: 1) training in simulation\tyghnm, or 2) training from real-world visual datasets~\cite{dasari2019robonet, mandlekar2018roboturk, jang2022bc, ebert2021bridge, nair2022r3m, chen2021learning, bahl2022human}.
This latter approach has been favored lately, even though the generalization obtained is restricted to that demonstrated in the datasets.
In long-horizon \textit{mobile manipulation}, however, the breadth of generalization demanded extends beyond objects. This, combined with the length and compositional variability of each activity (i.e., the same activity may require a different ordering of the same skills to be achieved), renders collecting a sufficiently broad distribution of real-world data less feasible. 
On the other hand, reaching the necessary generalization for mobile manipulation could be obtained from non-real-world trajectories.

When it comes to domain transfer, multiple solutions have been proposed for visual \textit{stationary manipulation} and \textit{navigation}, but they fall short when applied to mobile manipulation. The most common approach is to try to narrow the domain gap\abab. While successful in stationary manipulation and navigation, these methods may not be sufficient for long-horizon mobile manipulation that demands not only perceptual generalization but also compositionality in the solution.
Other methods have also achieved success in navigation and stationary manipulation by choosing input modalities that have lower domain gap \fdsfdsfd. While sufficing for navigation and some stationary manipulation, it is unclear if the input modalities chosen in these methods are sufficient for the fine-grained skills involved in many mobile manipulation activities. Finally, a family of adaptive learning\pepepe{} and system identification\okjnhgt{} algorithms have also achieved success in domain transfer in legged locomotion. While these methods narrow the domain gap in dynamics and action-state transition, it is not yet clear whether these methods can learn visuomotor solutions for a mobile manipulator.

To tackle long-horizon mobile manipulation solutions in the real world, we propose $\mobilemember$, or ``$\member$''-- a factorized method based on the $\ember$ framework~\cite{wu2021example}. Concretely, an activity is first factorized into a repertoire of primitive visuomotor skills, and $\member$ reinforcement-learns each skill in simulation, and transfers and recomposes these skills into a long-horizon mobile manipulation solution for the activity, achieving levels of robustness beyond what $\ember$ could do (see Sec.~\ref{sec:experiments}).
Thanks to the factorization of skills, $\member$ copes with initial and task conditions and is able to handle a mobile manipulation activity.

We demonstrated with extensive evaluations on a real-world mobile manipulator that $\member$ can complete a long-horizon activity (\texttt{cleaning\_kitchen}) with 53\% success by learning five different visuomotor skills in a simulator, concatenating them autonomously into sequences, and generalizing more robustly than existing mobile manipulation solutions.

\section{Related Work}
\subsection{Robot learning via domain transfer}
Domain transfer has a rich history in stationary robotic manipulation\tyghnm{}, navigation\ididid{}, and legged locomotion\sdfghjk{}. This includes the use of photorealism (e.g. photorealistic rendering\abababa{} or Generative Adversarial Networks\ababaab) and domain randomization\abab{} to narrow the domain gap. In comparison, $\member$ develops a factorized algorithm for long-horizon mobile manipulation, which demands both perceptual generalization as well as compositional generalization that can factorize and reuse learned visuomotor skills. Previous works also investigated the use of alternative observations\fdsfdsfd{} to narrow the domain gap. $\member$ accepts images as input and performs more fine-grained mobile manipulation than pick-and-place tasks. Finally, adaptive learning\pepepe{} and system identification\okjnhgt{} algorithms have achieved success in narrowing the dynamics sim-to-real gap. In comparison, $\member$ attempts to narrow the domain gap in mobile manipulation. 

\subsection{Learning mobile manipulation in simulation or real world}
Prior robot learning methods have achieved success in mobile manipulation in simulation\ikyhtg{} or the real world\rtyhuj. Some of these methods collect data in a continuous, online manner\ikjhuj{}, while others break data collection into primitives\iiuhnjmk{} to learn to perform long-horizon mobile manipulation. Inspired by these works, $\member$ performs long-horizon mobile manipulation, which demands perceptual generalization and compositional generalization that can factorize and reuse learned visuomotor skills.

\subsection{Reinforcement learning for stationary manipulation}
Reinforcement learning (RL) has a rich history of being used for robotic control from locomotion\lkjhg, navigation\ahi, to stationary manipulation \jkghkg. Indeed, prior methods in model-free\mfrl{} and model-based RL\mbrl{} achieved remarkable success in stationary manipulation. Drawing inspiration from these works, $\member$ extends $\ember$~\cite{wu2021example}, a previous work in this category, to performing long-horizon mobile manipulation.

\begin{figure*}[h]
\centering
\includegraphics[width=\textwidth]{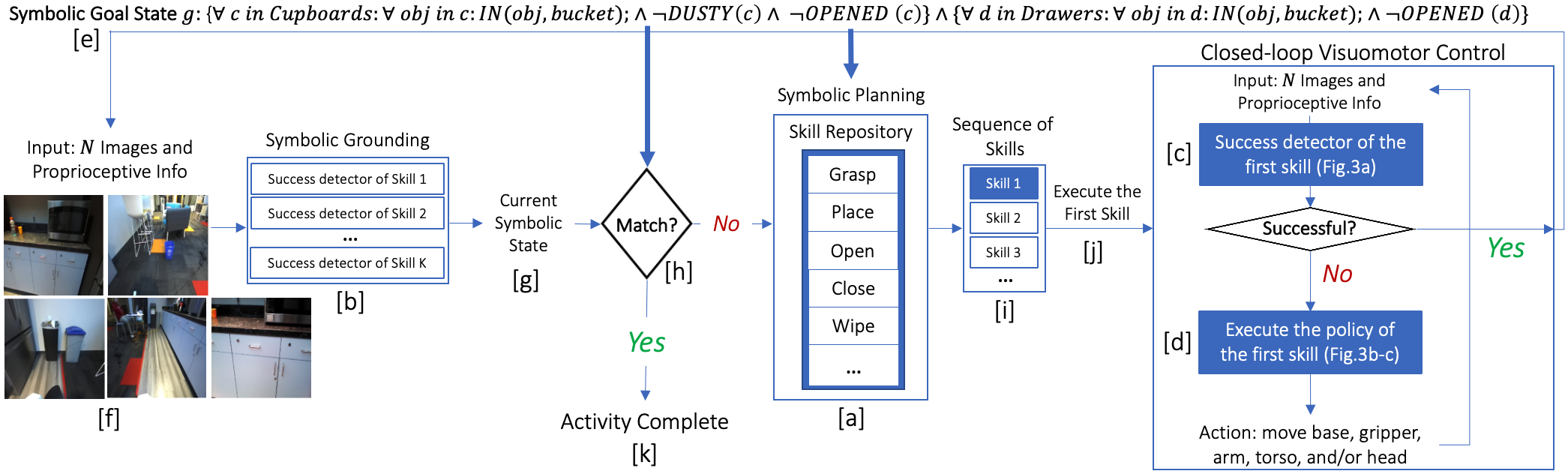}
\caption{\textbf{High-Level Overview of $\member$ for \texttt{cleaning\_kitchen} activity}. To begin, the human first specifies the activity using a symbolic goal state $g$ (Fig.~\ref{highlevel} [e]). For training, the activity is factorized into a repertoire of primitive skills [a] that $\member$ learns using RL in simulation along with per-skill success detectors [b,c]. In order to detect relevant objects in the environment, $\member$ also grounds the robot's raw pixel observations into a symbolic representation of the current state [b,g]. At test time, the per-skill success detectors are used to 1) map the input [f] --raw pixel observations ($N$ 112$\times$112 images) and proprioceptive information (joint angles)-- into a symbolic representation of the current state [g] to check full-activity completion [h] and perform long-horizon planning [a], and 2) during execution of a skill [d] to detect whether a skill has succeeded [c]. Long-horizon planning is performed by a symbolic planner [a] that computes the sequence of skills to perform [i] and executes the first skill of this sequence [j] based on both the goal [e] and current symbolic state [g]. This procedure repeats until the current symbolic state [g] matches the goal state [e], after which robot execution terminates successfully [k]. Modules [b,c,d] are learned modules elaborated in Sec.~\ref{s:mobilember} and Fig.~\ref{lowlevel}.}
\label{highlevel}
\end{figure*}
\section{Preliminaries}
\label{s:preliminaries}
\subsection{Modeling long-horizon mobile manipulation} 
In this work, we consider the problem of performing a long-horizon mobile manipulation activity: $\mathcal{M}$. We model the activity's environment as a controlled Markov process represented by the tuple $\mathcal{E} = \langle \mathcal{S}, \rho_0, \mathcal{A}, \mathcal{T}, \gamma, H \rangle$, with an observation space of $N$ cameras of resolution $H \times W$, which are $H=W=112$, and the robot's 19 joint angles, and thus $s \in \mathcal{S} = \mathcal{R}\{N \times 112 \times 112 \times 3 + 19\}$, an initial state distribution $\rho_0$, an action space $a \in \mathcal{A}$ (see Sec.~\ref{s:mobilember}), a dynamics model $\mathcal{T}: \mathcal{S} \times \mathcal{A} \times \mathcal{S} \to \mathbb{R}$, a discount factor $\gamma \in [0, 1)$, and a finite horizon $H$. We assume the goal of an activity is defined by a set of symbolic predicates in first-order logic that we obtain from BEHAVIOR~\cite{srivastava2022behavior,li2022behaviork}, a dataset of everyday activities defined in a domain-definition language (BDDL) similar to PDDL~\cite{mcdermott1998pddl}. For example, the BEHAVIOR \texttt{cleaning\_kitchen} activity is defined as: 

{\begin{align}
\nonumber&\{\forall \text{ cupboard} \in \text{cupboards: } \{\forall \text{ object} \in \text{cupboard:}\\\nonumber
&\text{\space\space\space\space (IN object bucket)\} $\land$ ($\neg$ OPENED cupboard) $\land$} \\\nonumber
& \text{\space\space\space\space ($\neg$ DUSTY cupboard)\} $\land$ } \{\forall \text{ drawer} \in \text{drawers:}\\\nonumber
&\{\forall \text{ object} \in \text{drawer: } \text{(IN object bucket)\} $\land$ }\\\nonumber
&\text{$\neg$ (OPENED drawer)\}}\nonumber
\end{align}}

In plain words, this means the goal is to relocate objects inside each cupboard and drawer in the environment into a bucket on the floor and ensure that cupboards and drawers are closed and that cupboards are not dusty. Performance of the robot in this activity is binary: ``success'' if the symbolic goal state is satisfied within a finite amount of real-clock time, or ``failed'' otherwise. Let $K$ denote the total number of unique mobile manipulation skills the robot has learned in simulation (in \ckck, $K=5$), and $k \in [1, K]$ denote the $k^{th}$ skill in the robot's skill repertoire. Here, each skill is a solution for a different Markov Decision Process (MDP) $\mathcal{M}^k = \langle \mathcal{E}, \mathcal{R}^k \rangle$, where the robot's environment $\mathcal{E}$ is shared across activities and skills, and $\mathcal{R}^k: \mathcal{S} \times \mathcal{A} \to \mathbb{R}$ is the reward function for the $k^{th}$ skill. This paper will use ``primitives'' vs. ``skills'' as well as ``factorize'' vs. ``decompose'' interchangeably.

\subsection{Factorization via \underline{E}xample-Driven \underline{M}odel-\underline{B}as\underline{E}d \underline{R}L ($\ember$)}
$\member$ is a factorized long-horizon mobile manipulation extension of $\ember$~\cite{wu2021example}. The goal of this work is to provide a preliminary solution to long-horizon mobile manipulation activities. 
To this end, $\ember$ is not enough: it is unable to cope with the variability and task length involved in mobile manipulation activities. Mobile manipulation may require a large amount of data to cover the activity distribution, and $\member$ overcomes this challenge by training mobile manipulation skills in simulation and applying them to a long-horizon mobile manipulation activity. 

\section{$\mobilemember$}
\label{s:mobilember}
$\mobilemember$, or ``$\member$'', is designed to overcome the limitations of $\ember$ and to be able to perform long-horizon mobile manipulation. Below, we first describe how a long-horizon mobile manipulation activity is factorized into a repository of visuomotor mobile manipulation skills during training, and recomposed to solve the long-horizon activity at test time. We then discuss how $\member$ learns each factorized visuomotor skill in simulation. Finally, we describe how $\member$ enables the factorized visuomotor skills to be used in the real world.

\subsection{Skill decomposition and recomposition of a long-horizon mobile manipulation activity}
To begin, the human first specifies the activity using a symbolic goal state $g$ (Fig.~\ref{highlevel} [e]). During training, the long-horizon mobile manipulation activity is factorized into a repository of skills based on the symbolic representation of this goal state. To verify that the visuomotor skills are successful, each skill is trained together with a ``success detector'' that will determine visually when the symbolic component of the activity goal has transitioned to the desired value. In order to detect relevant objects in the environment, $\member$ also grounds the mobile manipulation robot's observations into a symbolic representation of the current state (Fig.~\ref{highlevel} [b,g]). 

To perform the long-horizon activity at test time, $\member$ first computes the current symbolic condition of the environment by passing the visual observations (112$\times$112 images) and proprioceptive information (Fig.~\ref{highlevel} [f]) to the skills' learned success detectors (Fig.~\ref{highlevel} [b]). Using both the goal (Fig.~\ref{highlevel} [e]) and the current symbolic state (Fig.~\ref{highlevel} [g]), the symbolic planner computes the sequence of skills to perform (Fig.~\ref{highlevel} [i]) and executes the first skill of this sequence (Fig.~\ref{highlevel} [j]). This procedure repeats until the current symbolic state (Fig.~\ref{highlevel} [g]) matches the goal state perfectly (Fig.~\ref{highlevel} [e]), after which robot execution terminates successfully (Fig.~\ref{highlevel} [k]).

\begin{figure*}[h]
    \centering
    \includegraphics[width=\textwidth]{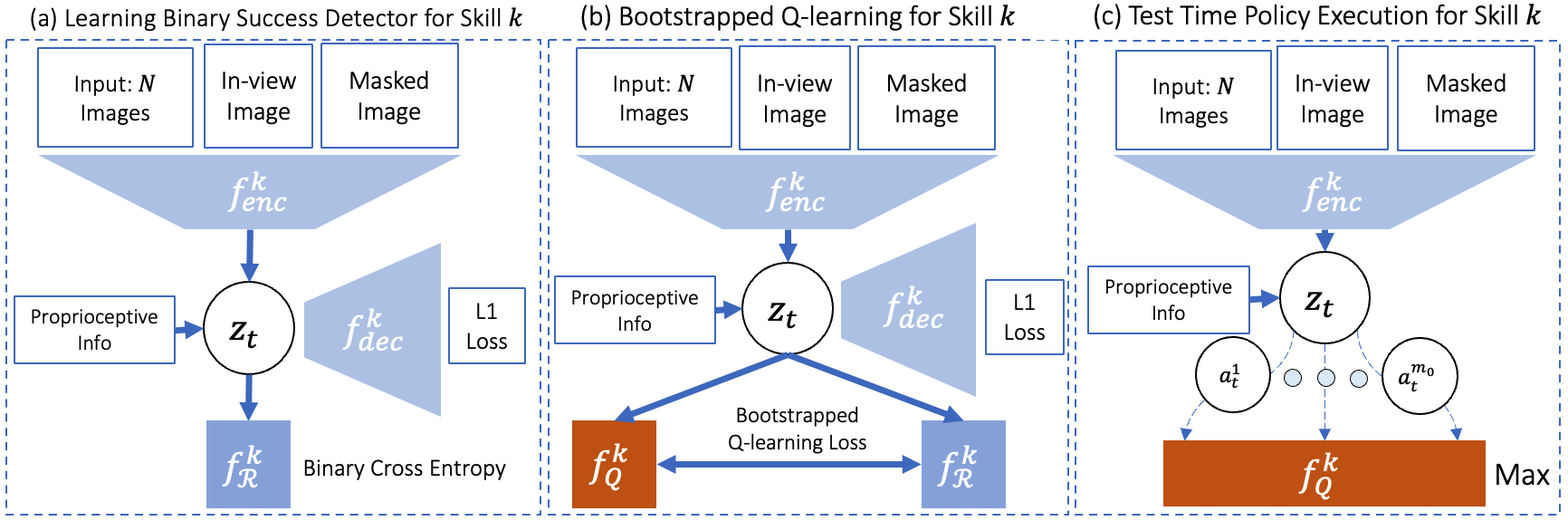}
    \caption{\textbf{Low-Level Visualization of $\member$'s Skill-Learning Process}. $\member$ learns to perform each skill by learning three individual components for each skill: a variational autoencoder (VAE) ($\vae$ in Fig.~\ref{lowlevel}), success detectors ($\rew$ in Fig.~\ref{lowlevel}), and Q-functions ($\q$ in Fig.~\ref{lowlevel}). The VAE reduces the dimensionality of the robot's pixel observation that is used as input for the success detectors and the Q-functions; the success detectors allow $\member$ to learn a binary reward function for each skill; and Q-functions are learned from the binary reward functions and allow $\member$ to perform visuomotor control. Kindly see the $\ember$ paper~\cite{wu2021example} for details.}
    \label{lowlevel}
\end{figure*}
\subsection{Learning Each Factorized Skill in Simulation}
$\member$ learns to perform each skill by learning three individual components per skill (see Fig.~\ref{lowlevel}): a variational autoencoder (VAE) ($\vae$ in Fig.~\ref{lowlevel}), success detectors ($\rew$ in Fig.~\ref{lowlevel}), and Q-functions ($\q$ in Fig.~\ref{lowlevel}). The VAE reduces the dimensionality of the robot's pixel observation into a latent representation that is used as input for the success detectors and the Q-functions; the success detectors allow $\member$ to learn a binary reward function for each skill; and Q-functions are learned from the binary reward functions and allow $\member$ to perform visuomotor control. Accordingly, the VAE optimization objective is: 
\begin{equation}
    \min_{\vae} \mathbb{E}_{s_t, a_t \sim \mathcal{D}}\bigg[ -\mathcal{L}_\text{vae} (s_t)\bigg]\nonumber
\end{equation}
where $\mathcal{D}$ is the dataset of RL trajectories, and $\mathcal{L}_\text{vae}$ is the evidence lower bound (ELBO) for the VAE: 
\begin{align}
\max_{p, q} \mathbb{E}_{q(z \mid s)} \left[\log p\left(s \mid z \right) \right] - D_\text{KL}{q(z \mid s)}{p(z)}\nonumber
\end{align}
The optimization objective for the success detector (Fig.~\ref{lowlevel} a) is
\begin{align}
    \max_{\rew} \mathbb{E}&_{s^+ \sim \mathcal{D}^+, z^+ \sim \enc(\cdot \mid s^+)}\left[ \log\left( \rew(z^+)  \right) \right] \nonumber\\
    & +\mathbb{E}_{s^- \sim \mathcal{D}, z^+ \sim \enc(\cdot \mid s^+)}\left[\log\left( 1 - \rew(z^-)  \right) \right]\nonumber
\end{align}
Here, $\mathcal{D}^+$ and $\mathcal{D}^-$ are the datasets of images labeled as positive and negative, $s^+$ and $s^-$ are the images sampled from $\mathcal{D}^+$ and $\mathcal{D}^-$. 
The Q-function optimization objective is:
\begin{align}
    \min_{{\q}} \mathbb{E}&_{s_t, a_t, s_{t+1} \sim \mathcal{D}}\nonumber\\
    & \left[\q \left(z_t, a_t \right)  - \left(\overline{\rew}\left(z_{t+1}\right) + \gamma \overline{\rew} \left( z_{t+1}\right) \qtarget\right)\right]^2\nonumber
\end{align}
where 
\begin{align}
\overline{\rew}(z) &\equiv \mathds{1}\{\rew(z) > 0.5\}\nonumber\\
z_t &\sim \vae(\cdot \mid s_t)\nonumber\\
z_{t+1} &\sim \vae(\cdot \mid s_{t+1})\nonumber
\end{align}
Finally, the Q-value target is:
\begin{align}
    \qtarget=\max_{a_{t+1}} \q(z_{t+1}, a_{t+1})\nonumber
\end{align}
It is computed by maximizing over 200 randomly and uniformly sampled actions $a^{1:200}_{t+1}$. There are three significant differences between $\member$ and $\ember$: 1) while $\ember$ learns and rolls out a latent dynamics model for MBOLD~\cite{tian2020model} planning, $\member$ does not learn or use a latent dynamics model given its poor empirical performance in the real world, which we quantify in Sec.~\ref{sec:experiments}. Instead, we directly apply cross-entropy method (CEM) on actions sampled from the action space and Q-values queried from the learned Q-functions~\cite{kalashnikov2018qt,Kalashnikov2021MTOptCM}. 
2) In addition to per-camera images, $\member$ takes as input the image in which the object resides in (``In-view Image'' in Fig.~\ref{lowlevel}) as well as the masked image (``Masked Image'' in Fig.~\ref{lowlevel}) from the object of interest. These two additional images signal to the learned $\member$ model which camera the object of interest resides in. 3) The VAE in $\member$ is per-skill instead of shared across skills as in $\ember$.

\subsection{Training learned visuomotor skills to be used in real-world conditions}
To allow learned visuomotor skills to be used in real-world conditions, we use photorealism and domain randomization in a simulator built on top of iGibson~\cite{xia2018gibson,shen2021igibson,li2021igibson}. To execute actions in the real world, $\member$ commands small changes in the pose of the robot's two end-effectors (6D) and base (3D), and joint positions of the head, torso and two grippers. 
\begin{figure*}
\centering
\includegraphics[width=\textwidth]{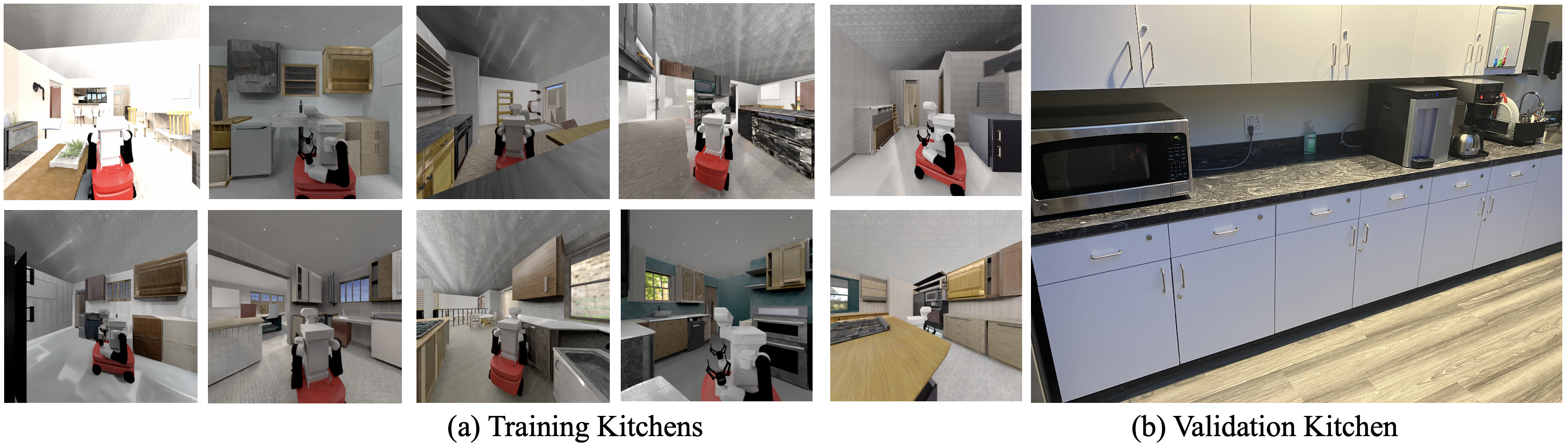}
\caption{\textbf{Train and Validation Environments.} Fig.~\textbf{(a)} exhibits a subset of simulated kitchen environments in which the TIAGo robot learns each mobile manipulation primitive skill. The TIAGo robot is equipped with two Robotiq parallel-jaw grippers and cameras capturing images at 3Hz. To narrow the domain gap, we use a real world kitchen for validation (Fig.~\textbf{(b)}), and then evaluate $\member$ on three kitchens.}
\label{fig:envs}
\end{figure*}
~\label{sec:simulation}

\subsection{Photorealism and domain randomization}
We apply photorealistic rendering to the iGibson simulator~\cite{xia2018gibson,shen2021igibson,li2021igibson} to increase the photorealism of the simulated scenes. We then randomize rendered scenes across 14 dimensions: 
\begin{itemize}
\item \textit{Object instance:} each object is randomized across instances within the same category
\item \textit{Object placement:} when learning each skill, each object is placed in the environment with a randomized 6-DOF pose
\item \textit{Object scale:} each object in the environment is randomized across dimensions and scales
\item \textit{Object texture:} texture is randomized across object texture maps
\item \textit{Indoor lighting condition:} lighting direction, types, surface area, and intensity are randomized in the indoor environment 
\item \textit{Outdoor lighting condition:} dome lighting direction and intensity are randomized
\item \textit{Initial robot placement:} the robot is initially randomly (2-DOF position and 1-DOF rotation) placed in the kitchen
\item \textit{Initial camera viewpoint:} the initial viewpoint of the camera is randomized by the starting joint configuration of the robot, so long as the objects are still visually accessible by at least one of the robot cameras
\item \textit{Camera parameters:} each training environment is randomized across intrinsic and extrinsic camera parameters
\item \textit{Outdoor environment texture:} outdoor texture is randomized across HDR environment maps
\item \textit{Indoor interior randomization:} ceilings, walls, and floors are randomized across textures maps
\item \textit{Scene randomization:} each training environment is randomized across rooms and floor plans
\item \textit{Physics:} each training environment is randomized across 0.5-3.5x frictional and inertial coefficients
\item \textit{Robot arm texture:} each training environment is randomized across 20-25 robot arm textures
\end{itemize}

\section{Experiments}
\label{sec:experiments}

Experiments in this paper aim to answer four main questions: 1) Can $\member$'s factorized and learned primitive skills generalize? 2) Does latent dynamics prediction in $\ember$ contribute or degrade $\member$'s performance? 3) Can $\member$ perform long-horizon activities? 4) How much do photorealism and domain randomization each contribute to $\member$'s performance?

\begin{figure*}
    \centering
    \resizebox{\textwidth}{!}{
    \subfloat[][Validation (L), Test (R) Buckets]{
    \includegraphics[width=0.35\textwidth,trim={0 10 0 30},clip]{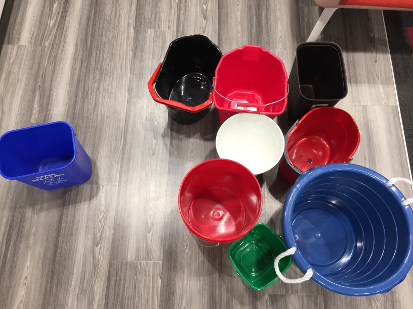}}
    \subfloat[][Validation (L), Test (R) Object Instances]{
    \includegraphics[width=0.37\textwidth,trim={0 10 0 30},clip]{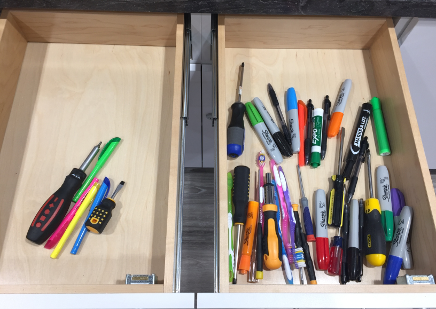}}
    \subfloat[][Validation (L), Test (R) Clothes]{
    \includegraphics[width=0.33\textwidth,trim={0 0 0 200},clip]{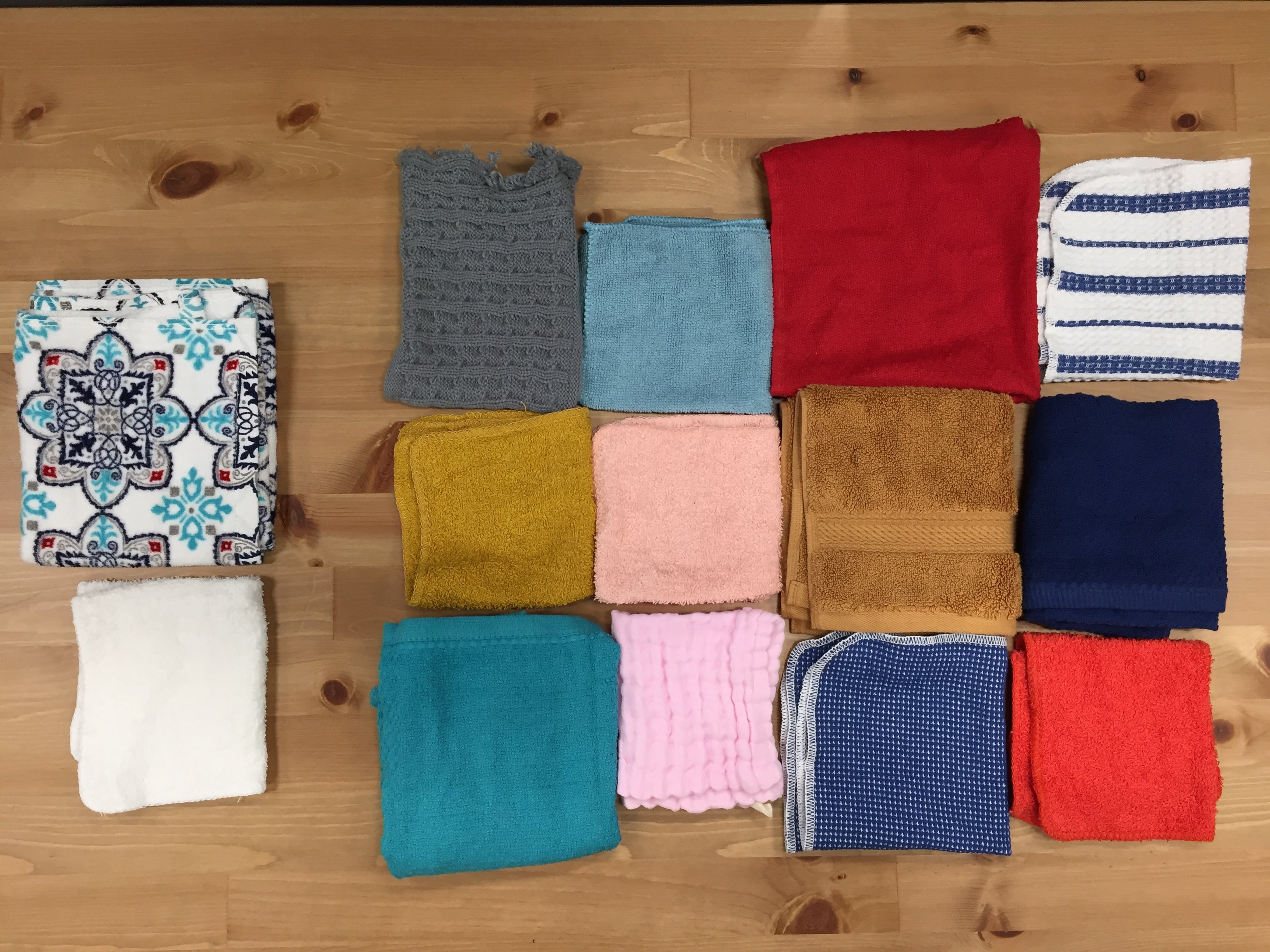}}}
    \caption{\small Validation and Test buckets, object instances, and wiping clothes used in real-robot experiments}
    \label{fig:objects}
\end{figure*}
\begin{table*}[t]
\centering
\caption{\small Successful trials (out of 30) and percentages of \ckck{} activity. $K$: number of skills composed in each trial.} 
\label{tab:long}
\small
\resizebox{\linewidth}{!}{
\begin{tabular}{c|c|c|c|c|c|c|c}
\toprule
\multirow{3}{*}{\ckck} & \multirow{2}{*}{$\member$} & Prior Method 1: & Prior Method 2: & Prior Method 3: & Ablation 1: & Ablation 2: \\
& \multirow{2}{*}{(Ours)} & $\member$ w/ & $\ember$ ($\member$ w/ & BEE ($\member$ w/ & $\member$ w/o & $\member$ w/ 50\% \\ 
& & Scripted Skills & MBOLD planning)~\cite{wu2021example} & Visual Foresight)~\cite{chen2020batch} & Photorealism & Domain Randomization \\\midrule
Kitchen 1 & 5/10 & 0/10 & 0/10 & 0/10 & 0/10 & 2/10 \\ 
Kitchen 2 & 5/10 & 1/10 & 0/10 & 0/10 & 0/10 & 2/10\\
Kitchen 3 & 6/10 & 0/10 & 0/10 & 0/10 & 0/10 & 3/10\\\midrule
Total & 16/30 (53.3\%)  & 1/30 (3.3\%) & 0/30 (0\%) & 0/30 (0\%) & 0/30 (0\%) & 7/30 (23.3\%) \\
\bottomrule
\end{tabular}}
\end{table*}

\subsection{Experimental setup}
To answer these questions, we conduct experiments of both factorized primitive skills and the long-horizon \ckck{} BEHAVIOR~\cite{srivastava2022behavior} activity across three kitchens and dozens of object instances (Fig.~\ref{fig:objects}). In Fig.~\ref{fig:envs}, TIAGo, a bi-manual mobile manipulator, has access to cameras capturing 112$ \times$112 images at 3Hz, as well as 360$\degree$ 2D LiDAR scans.

\subsection{Comparisons}
We compare $\member$ to two prior methods: $\ember$~\cite{wu2021example} ($\member$ with MBOLD~\cite{tian2020model} planning) and BEE~\cite{chen2020batch} ($\member$ with Visual Foresight, which uses success detectors instead of the Q-functions for model predictive control). We also compare $\member$ to ``$\member$ with scripted skills'', which scripts instead of learning skills. The training data for experiments are collected from the same simulator, which contains photorealism and domain randomization techniques outlined in Section~\ref{sec:simulation}. 

\subsection{Factorized skill performance}
We compare all methods across five factorized skills: grasp, place, open, close, and wipe. $\member$ achieves higher success rates across three kitchens and 50+ object instances (Fig.~\ref{fig:objects}). In contrast, $\ember$ and BEE achieve lower success rates for each skill due to the difficulty of using latent dynamics models trained from simulation in the real world.

\subsection{Long-horizon activity performance}
For long-horizon experiments, this paper investigates the \ckck{} activity, in which the robot is placed in a kitchen, which contains cupboards or drawers. Each cupboard or drawer is closed initially, and there are object instances placed in it. There is a bucket~(Fig.~\ref{fig:objects}) randomly placed on the floor. There is a piece of wiping cloth~(Fig.~\ref{fig:objects}) laying in each cupboard for the robot to wipe the cupboard shelf clean. After wiping, the cloth should also be placed in the bucket. The goal of this activity is to put objects in each cupboard and drawer into the bucket, wipe cupboards clean, and close cupboards and drawers. Wiping is “clean” if the majority of the surface area of the cupboard shelf reachable by the robot is wiped.

In Table~\ref{tab:long}, we find that ``$\member$ (Ours)'' completes the activity with 53.3\% success. In comparison, ``$\member$ with Scripted Skills'' achieves single-digit success rates due to the use of scripted, non-learning skills, while $\ember$ and BEE achieve no success due to poor performance of the latent dynamics models in the real world. Empirically, we observe three major failure cases of $\member$ in this activity: 1) collision with the kitchen cupboards or drawers; 2) leaving at least one object inside a cupboard or drawer; 3) failed object grasp attempts that let to objects becoming no longer reachable.

\subsection{Ablation studies}
We conduct ablations to quantify the performance contribution of photorealism and domain randomization. To quantify the contribution of photorealism, we ablate $\member$ by turning off photorealism in simulation. In Table~\ref{tab:long}, we find that $\member$'s long-horizon mobile manipulation activity success rates degrade by 53.3\%, quantifying the importance of photorealistic rendering in $\member$'s performance.
To ablate domain randomization, we shrink the range of randomization across each dimension of randomization specified in Section~\ref{sec:simulation} by 50\%. In Table~\ref{tab:long}, we find that $\member$'s long-horizon mobile manipulation activity success rates degrade by 30\%, quantifying the importance of rendering randomized environments and objects to $\member$'s performance.

\section{Conclusion}
In this work, $\member$ develops a factorized algorithm for long-horizon mobile manipulation by factorizing and reusing learned skills. Nevertheless, the domain gap is immense, which is not only reflected in the low success rates achieved in experiments but also in the limited settings and activities the $\member$ algorithm can handle. We hope that this work demonstrates the difficulty of mobile manipulation as an open research problem.

\section{Appendix: Additional Implementation Details and Clarifications}
\subsubsection{Using Raw Pixels vs. Depth or RGB-D as Visual Observations}
While $\member$ uses images as input due to legacy considerations, it is a limitation that it currently cannot accept depth input (inclusively or exclusively) as part of its visual observations. Such important extension from raw pixels to depth image is left for future work.
\subsubsection{Proprioceptive Information as Non-visual Observations}
In additional images, $\member$ takes as input proprioceptive information as non-visual observations. Such proprioceptive information comprises robot joint angles (e.g. head, torso, arms, and grippers).
\subsubsection{Use of $360\degree$ 2D LiDAR}
The $360\degree$ 2D LiDAR readings of the robot provide fail-safe mechanisms for the robot and are not visible to the learning modules in $\member$.
\subsubsection{Simulation Environments}
The simulation environments used in this paper are those produced from the iGibson simulator.
\subsubsection{Using additional photorealistic datasets}
While iGibson is used as the primary simulator in this paper, extending training environments to other widely used ones such as Matterport, Replica as well as Habitat (including the Rearrangement Challenge) is respectfully left for future work. The authors agree to some extent that submissions to this challenge are probably more realistic baselines for the chosen formulation.
\subsubsection{The stringent demands of generalization abilities of solutions to the mobile manipulation problem}
The authors would like to clarify that the combinatorial nature of the mobile manipulation problem is what makes the demands of the generalization abilities of mobile manipulation solutions particularly stringent. 
\subsubsection{On the use of ``unclear'' in discussions of related works}
The authors acknowledge that instead of conjecturing that it is ``unclear'' the existing methods will work, they could have explained how they empirically validate these concerns and propose their approach as a solution. However, given the empirical weakness of their own solutions, the authors have opted out of adopting a stronger connotation at this time. 
\subsubsection{Mobile vs. stationary manipulation}
The authors acknowledge that the paper might come off as suggesting that mobile manipulation is a
totally different problem than stationary manipulation. However, the authors would like to clarify that they do not think mobile manipulation is a fundamentally different problem from stationary manipulation.

\section{Acknowledgments}
The authors would first like to give their special thanks to William Chong, Marion Lepert, and Wesley Guo from the Department of Mechanical Engineering of Stanford University for their outstanding mechanical support for the robot. The authors would also like to thank the entire Stanford Vision and Learning Lab (SVL) for its wonderful support, such as experimental setup, hardware support, and computational resources.

\renewcommand*{\bibfont}{\footnotesize}
\printbibliography

\end{document}